\documentclass[lettersize,journal]{IEEEtran}
\usepackage{amsmath,amsfonts}
\usepackage{algorithmic}
\usepackage{algorithm}
\usepackage{array}
\usepackage[caption=false,font=normalsize,labelfont=sf,textfont=sf]{subfig}
\usepackage{textcomp}
\usepackage{stfloats}
\usepackage{url}
\usepackage{verbatim}
\usepackage{graphicx}
\usepackage{cite}
\usepackage{makecell}
\usepackage{multirow}
\usepackage{threeparttable}
\begin{document}

\title{LaSNN: Layer-wise ANN-to-SNN Distillation for Effective and Efficient Training in Deep Spiking Neural Networks}

\author{
	Di~Hong,
	Jiangrong Shen,
	Yu~Qi$^{*}$,
	Yueming Wang$^{*}$ 
\thanks{This work was partly supported by the grants from National Key R\&D Program of China (2018YFA0701400), Key R\&D Program of Zhejiang (2022C03011), Natural Science Foundation of China (61906166, 62276228), the Fundamental Research Funds for the Central Universities, the Starry Night Science Fund of Zhejiang University Shanghai Institute for Advanced Study (SN-ZJU-SIAS-002), and the Lingang Laboratory (LG-QS-202202-04).
	
	Di Hong is with the College of Computer Science and Technology, and the Qiushi Academy for Advanced Studies, Zhejiang University, Hangzhou, China. Jiangrong Shen is with the College of Computer Science and Technology, Zhejiang University, Hangzhou, China. Yu Qi is with the MOE Frontier Science Center for Brain Science and Brain-machine Integration, and the Affiliated Mental Health Center \& Hangzhou Seventh People's Hospital, Zhejiang University School of Medicine, Hangzhou, China. Yueming Wang is with the Qiushi Academy for Advanced Studies, Zhejiang University, Hangzhou, China.
	The corresponding authors are Yu Qi (qiyu@zju.edu.cn) and Yueming Wang (ymingwang@zju.edu.cn).
}}



\maketitle

\begin{abstract}
	Spiking Neural Networks (SNNs) are biologically realistic and practically promising in low-power computation because of their event-driven mechanism. Usually, the training of SNNs suffers accuracy loss on various tasks, yielding an inferior performance compared with ANNs. 
	A conversion scheme is proposed to obtain competitive accuracy by mapping trained ANNs' parameters to SNNs with the same structures. 
	However, an enormous number of time steps are required for these converted SNNs, thus losing the energy-efficient benefit. 
	Utilizing both the accuracy advantages of ANNs and the computing efficiency of SNNs, a novel SNN training framework is proposed, namely layer-wise ANN-to-SNN knowledge distillation (LaSNN).
	In order to achieve competitive accuracy and reduced inference latency, LaSNN transfers the learning from a well-trained ANN to a small SNN by distilling the knowledge other than converting the parameters of ANN.
	The information gap between heterogeneous ANN and SNN is bridged by introducing the attention scheme, the knowledge in an ANN is effectively compressed and then efficiently transferred by utilizing our layer-wise distillation paradigm. 
	We conduct detailed experiments to demonstrate the effectiveness, efficacy, and scalability of LaSNN on three benchmark data sets (CIFAR-10, CIFAR-100, and Tiny ImageNet). 
	We achieve competitive top-1 accuracy compared to ANNs and 20x faster inference than converted SNNs with similar performance. 
	More importantly, LaSNN is dexterous and extensible that can be effortlessly developed for SNNs with different architectures/depths and input encoding methods, contributing to their potential development.
\end{abstract}


\section{Introduction}
	\IEEEPARstart{S}{piking} Neural Networks (SNNs) have been proven hopefully in lower-power computing, generally termed the third generation of neural networks \cite{Ghosh-Dastidar2009}. 
	Inspired by the principles of brain computing, SNNs imitate the brain that utilizes discrete spikes for representing and transmitting information, which provides significant potential on event-driven and energy-efficient neuromorphic hardware \cite{TrueNorth, navaridas2015spinnaker, zou2020hybrid}. 	
	One critical challenge of SNNs is the effective training of the spiking neuron-based networks \cite{tavanaei2019deep}. 
	Incorporating inspirations from the biological nervous systems and structures, one way to address the problem is by adopting the spike-time-dependent plasticity (STDP) learning rule, a specialization of the widely studied Hebbian rule in neuroscience. 
	However, lacking global instructors results in SNNs trained with the STDP rule cannot be extended to deep networks with good performances and are constrained to shallow structures for solving simple tasks (e.g., MNIST) \cite{diehl2015unsupervised, diehl2015fast, kheradpisheh2018stdp}.
	Due to the inherent non-differentiable problem in SNNs, the well-known backpropagation (BP) algorithm \cite{david1986learning} is not straightforwardly applicable to SNNs. 
	In general, replace the actual gradient in SNNs with a designed continuous function to approximate the discontinuous derivative of spiking neurons \cite{wu2019direct, NeftciMZ19}. 
	This approximation is adequate for small SNNs, but ineffective if SNNs adopt deep structures as well as solve challenging tasks (e.g., ImageNet). 	
	In addition, SNNs trained with surrogate-gradient methods fail to achieve comparable performance as their deep neural networks counterpart (ANNs). 
	In order to address the performance problem mentioned above, ANN-to-SNN conversion methods are developed that convert the parameters of pre-trained ANNs into SNNs with the same structures \cite{CaoCK15, Sengupta2019, han2020rmp, deng2021optimal}.
	Although these methods achieve nearly equivalent representation even in large networks, the ANN-converted SNNs usually need an enormous number of time steps (hundreds or even thousands) \cite{Sengupta2019, CaoCK15}, leading to intense power consumption. 
	Moreover, binary computation cannot be accelerated by using GPU devices, leading to high latency ($T$ $\times$ more time than ANN training). 
	These obstructions run contrary to the objective of low-power computing.
	
	Then it is natural to ask: is there an optimal way that both leverage the guidance of well-trained ANNs (like the conversion scheme) while simultaneously maintaining the efficiency of spike-based computing (like the surrogate scheme)? To this end, by distilling the knowledge from ANNs with a layer-wise scheme, we develop a novel framework called LaSNN. 
	Similar to conversion methods, utilizing a well-trained ANN to guide the SNN training process. However, instead of directly converting the network parameters, we propose to \textit{distill the knowledge} from an ANN to target SNN for low-power and fast inference. 
	
	To achieve this, the critical issue is how to distill knowledge from ANNs to SNNs effectively. 	Knowledge distillation generally transfers learning from a cumbersome teacher model with high performance to a smaller student model. Although knowledge distillation is promising for improvements in classification accuracy and model compressing, its application is mainly constrained to homogeneous models (e.g., from ANN to ANN) \cite{Hinton2015, ZagoruykoK17}. The spike-based representation and computation for SNNs differ highly from ANNs that use analog values. Therefore, previous distillation approaches and training techniques cannot be directly applied to heterogeneous models.
	
	In order to address these challenges, a new framework is developed, enabling SNNs to learn from ANNs. For clarity, we summarize the significance of our contributions as follows.
	\begin{enumerate}
		\item Instead of directly converting the network parameters from ANNs to SNNs, we propose an ANN-to-SNN knowledge transfer approach, which uses attention as a shared information representation, to bridge the information gap between heterogeneous networks of ANN and SNN.
		
		\item We put forward a layer-wise ANN-to-SNN distillation framework capable of transferring the knowledge from ANNs to SNNs. The training pipeline of LaSNN consists of three stages. Firstly, training a cumbersome ANN as the teacher model and, secondly, utilizing ANN-to-SNN conversion methods to initialize the parameters of the student SNN. Thirdly, training initial SNN based on a layer-wise distillation scheme that the student SNN is encouraged to imitate the inference of the teacher ANN. The proposed framework enables SNNs to compress the knowledge from a large ANN for achieving accurate and efficient computing.
		
		\item Detailed experiments are conducted on CIFAR-10, CIFAR-100, and Tiny ImageNet data sets to evaluate our methods. Experimental results illustrate that LaSNN can achieve compatible top-1 accuracy to ANNs and are 20x faster than converted SNNs with similar accuracy. More importantly, LaSNN is feasible for different architectures/depths and encoding methods. Our work, thus, strongly suggests the superiority of LaSNN for training deep SNNs.
	\end{enumerate}

\section{Related Work}
	Various training algorithms for SNNs have been developed that can be broadly categorized into synaptic plasticity learning rules, surrogate-gradient training methods, and ANN-to-SNN conversion algorithms. 
	Based on the sensitivity of time and abstracted the learning rules of biological brains, synaptic plasticity algorithms update the weights of connection according to neurons' firing time interval \cite{diehl2015unsupervised, kheradpisheh2018stdp, taherkhani2018supervised}. 
	Because of lacking global information, synaptic plasticity algorithms are typically applied in solving simple tasks and processing neuromorphic images \cite{diehl2015unsupervised, tavanaei2016bio, amir2017low}. 
	Therefore, our discussions are mainly focused on surrogate-gradient training and ANN-to-SNN conversion algorithms that have achieved significant progress on complex tasks and deep structures. 
	In addition, we briefly introduce previous works on knowledge distillation, which refers to transferring the knowledge in a large model or an ensemble of models into a single model that is much easier to deploy, and its application in SNNs.
	
	\subsection{Training of SNNs}	
		\subsubsection{Surrogate-gradient Algorithms}
			ANNs have obtained significant achievements with the gradient-based training algorithm. By computing the gradient of each operation during propagating forward, errors can backpropagate from the output layer to the input layer. 
			However, the derivatives of spike functions, such as integrate-and-fire (IF) and leaky-integrate-and-fire (LIF) neurons, are undefined at the time step of generating spikes and '0' otherwise. 
			In order to cope with the non-differentiable spiking computations, surrogate gradient methods use a continuous function to approximate the IF neuron, serving as a surrogate for the actual gradient. 
			Thus, SNNs can be directly optimized by applying backpropagation through time (BPTT) that acts like RNN \cite{wu2019direct, LeeKDSNN} or error backpropagation algorithms. 
			However, backpropagation through time algorithm needs to compute the gradient and backpropagate errors through each time step, thus are computationally expensive and slow. 
			In addition, the inaccurate approximations for computing the gradients accumulate errors failing to train deep SNNs directly.
		\subsubsection{ANN-to-SNN Conversion}
			Converting ANNs directly into SNNs has been proven successful for training deep SNNs \cite{CaoCK15, diehl2015fast, wu2019direct, Sengupta2019, han2020rmp, deng2021optimal}. 
			In order to facilitate the nearly lossless conversion, an ANN is constructed with rectified linear unit (ReLU) neurons and some restrictions (e.g., no bias terms, average polling, and no batch normalization), then trained with gradient descent. 
			Different mapping strategies are applied for converting the well-trained ANN to an SNN with IF neurons, such as data-based normalization \cite{diehl2015fast}, threshold balancing \cite{Sengupta2019, diehl2016truehappiness}, and soft reset (also called reset-by-subtraction mechanism) \cite{CaoCK15}. 
			Since only introducing a large number of time steps can make the firing rates closely approximate the high-precision activation, the major bottleneck of these methods suffers from high inference latency (2000-2500 time steps) to obtain satisfying performance. 
			In recent work, the authors dedicatedly calibrate the parameters in the SNN layer-by-layer to match the activation after conversion \cite{li2021free}. 
			However, it requires cumbersome architectures and is not scalable for small SNNs.		
		\subsubsection{Hybrid SNN Training}
			Recently, a hybrid training mechanism, including backpropagation and ANN-to-SNN conversion, has been introduced to overcome the intense computing of BPTT on the one hand and maintain the low latency of inference (100-250 time steps) on the other hand \cite{RathiSP020}. 
			Specifically, SNNs are converted from pre-trained ANNs, followed by fine-tuning with a surrogate gradient or BPTT. However, it still employs inherent information to optimize the weights of the SNN and lacks additional guidance. 
			Therefore, we need to optimize the balance between accuracy and latency further.  

	\subsection{Knowledge Distillation}
		Knowledge distillation has proven to be a practical approach to compress models by transferring the knowledge of a cumbersome model (teacher) to a small model (student). 
		Generally, soft labels that contain more information than one-hot labels are employed in the cross-entropy loss function to regularize the student model \cite{Hinton2015}. 
		Previous works mainly focus on ANN-to-ANN distillation by reformulating the supervision signals for more effective knowledge transfer, such as attention transfer \cite{ZagoruykoK17}. 

	\subsection{SNN Training with Knowledge Distillation}
		Previous works have shown that knowledge distillation approaches proposed in ANNs can extend with SNNs, such as distilling spikes from a large SNN to a small SNN \cite{KushawahaKBV20} and distilling label-based knowledge from a well-trained ANN to a simple SNN \cite{TakuyaZN21, LeeKDSNN}. 
		However, the approaches mentioned above mainly utilize label-based information, so the performance of deep SNNs can be improved further. 
		
	In this article, we first introduce the attention-based distillation capable of bridging the information representation and transmission gap between ANNs and SNNs. Based on the attention scheme, we propose a layer-wise distillation framework called LaSNN that effectively and efficiently transfers the learning of models. Moreover, we apply the surrogate-gradient method to optimize the parameters of SNNs during knowledge distilling.

\section{The LaSNN Framework}
	\begin{figure*} [tb]
		\centering
		\includegraphics[width=0.95\textwidth]{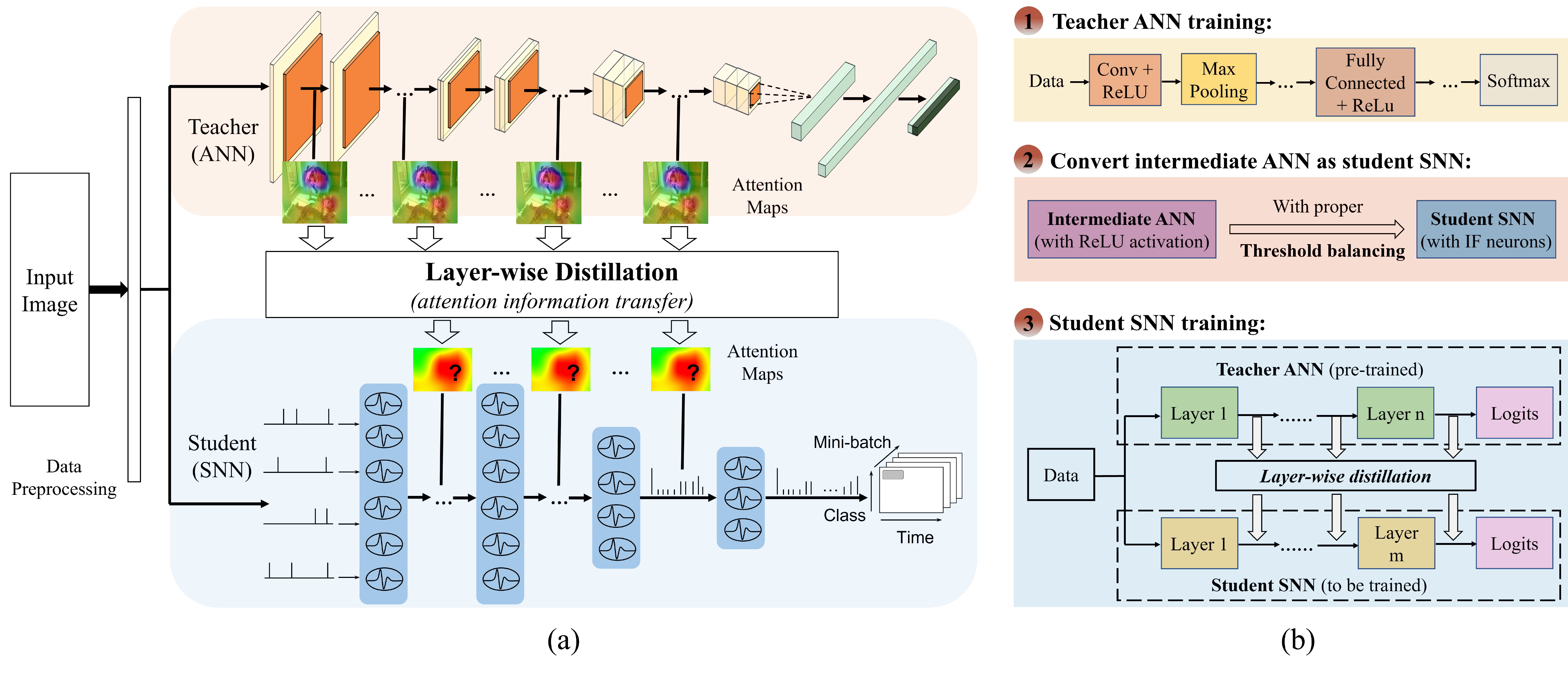}
		\caption{(a) The framework of the proposed LaSNN. (b) Training pipeline. }
		\label{fig:framework}
	\end{figure*}

	The overall process of layer-wise ANN-to-SNN knowledge distillation is illustrated in Fig. \ref{fig:framework} (a). 
	In this section, we first describe the SNN model. Then we give details about the ANN-to-SNN distillation paradigm with a layer-wise supervision strategy. Finally, we present the three-stage training process of LaSNN step-by-step.

	\subsection{SNN Model}
		\subsubsection{Encoding Method}
		We encode input images (analog values) into spatial-temporal patterns (spike trains) with a temporal coding scheme since the representation and transmission of information in SNNs are spikes. 
		The pixel values of input images are normalized to the range of $[-1, 1]$. Feeding these normalized values into the Poisson encoder outputs Poisson spike trains with rates proportional to the intensity of corresponding pixels. 
		Specifically, after inputting the image, the Poisson encoder will generate random numbers at every time step for every pixel, then compare these random numbers with the normalized value, outputting spikes if these random numbers are less than the normalized value. 
		Therefore, averaged over a long time, these Poisson-distributed spike trains are equivalent to pixel values. 
		
		\subsubsection{Spiking Neuron}
			In this work, we use LIF spiking neuron model. The iterative model in a discrete manner for machine learning is described by: 		
			\begin{equation}
				\label{LIF_iter}
				v_{i}^{t} =\lambda v_{i}^{t-1}+ \sum_{j}\omega_{ij}o_{j}^{t}-\theta o_{i}^{t-1} 
			\end{equation}
			\begin{equation}
				\label{LIF_binary}
				o_{i}^{t-1} =
				\left\{\begin{matrix}
					1,& if\ v_{i}^{t-1}>\theta \\
					0,& otherwise \\
				\end{matrix}\right.
			\end{equation}
			where $o$ denotes the spike output, $v$ represents the membrane potential, subscript $i$ and $j$ denote the post- and pre-neuron, respectively, superscript $t$ is the time step, $\omega_{ij}$ is the synaptic weight connecting the post- and pre-neuron, $\theta$ is the threshold potential, and $\lambda (< 1)$ indicates the leak in membrane potential. 
			In order to reduce the trainable parameters, neurons within the same layer share the identical threshold value, and all neurons share the same leak value. 
			In the process of forward propagation, the membrane potential of the neuron will increase by receiving pre-synaptic spikes, after reaching the threshold of firing, the neuron outputs a post-synaptic spike, and the membrane potential resets to resting. 
			
			However, as shown in equation (\ref{LIF_output}), in order to define the loss function on spike count, the neuron dynamics in the output layer are modified by removing the leaking part ($\lambda = 1$) and integrating the input without firing. 
			Neurons' number in the output layer is consistent with the number of classification targets, and the output predicted distribution $p$ is given in equation (\ref{output_prob}), 
			\begin{equation}
				\label{LIF_output}
				v_{i}^{t} = v_{i}^{t-1}+ \sum_{j}\omega_{ij}o_{j}^{t}
			\end{equation}
			\begin{equation}
				\label{output_prob}
				p_{i} = \frac{e^{v_{i}^{T}}}{\sum_{j=1}^{N}e^{v_{j}^{T}}}
			\end{equation}
			$T$ denotes the total number of time steps, $v^{T}$ represents neuron's membrane potential in the output layer over all time steps, and $N$ denotes the number of classification targets. 

		\subsubsection{Network Architectures}
			The SNN architecture is basically similar to traditional multi-layer feed-forward deep neural networks, such as VGG and residual architecture. 
			However, some details are modified to achieve a minimal loss in ANN-to-SNN conversion. 
			
			Firstly, since adopting bias terms increases the difficulty of threshold balancing and the probability of conversion loss, no bias terms will be used. Batch Normalization in ANN is also unused because eliminating the bias term results in the input bias of each layer becoming zero.  Dropout \cite{srivastava2014dropout} is employed for both ANN and SNN as an alternative regularizer. 
			
			Secondly, we adopt the average pooling operation to reduce the size of the feature map. Introducing max pooling operation results in significant information loss because of the binary activation in SNNs. 
			
			Thirdly, we replace the original wide kernel (7x7, stride 2) in residual architectures with an alternative block that consists of three small convolution layers (3x3, stride 1) and two dropout layers in between.			 
			
	\subsection{ANN-to-SNN Distillation}
		In order to distill the knowledge from an ANN to target SNN. The teacher is a well-trained cumbersome ANN model, containing rich and accurate attention information. The student is a small SNN model with similar ANN architecture. We explain 1) the activation-based scheme and 2) the gradient-based scheme for representing attention information, which bridges the knowledge gap between a real-valued ANN and a discrete signal SNN. 
		
		\begin{figure}[h]
			\centering
			\includegraphics[width=0.4\textwidth]{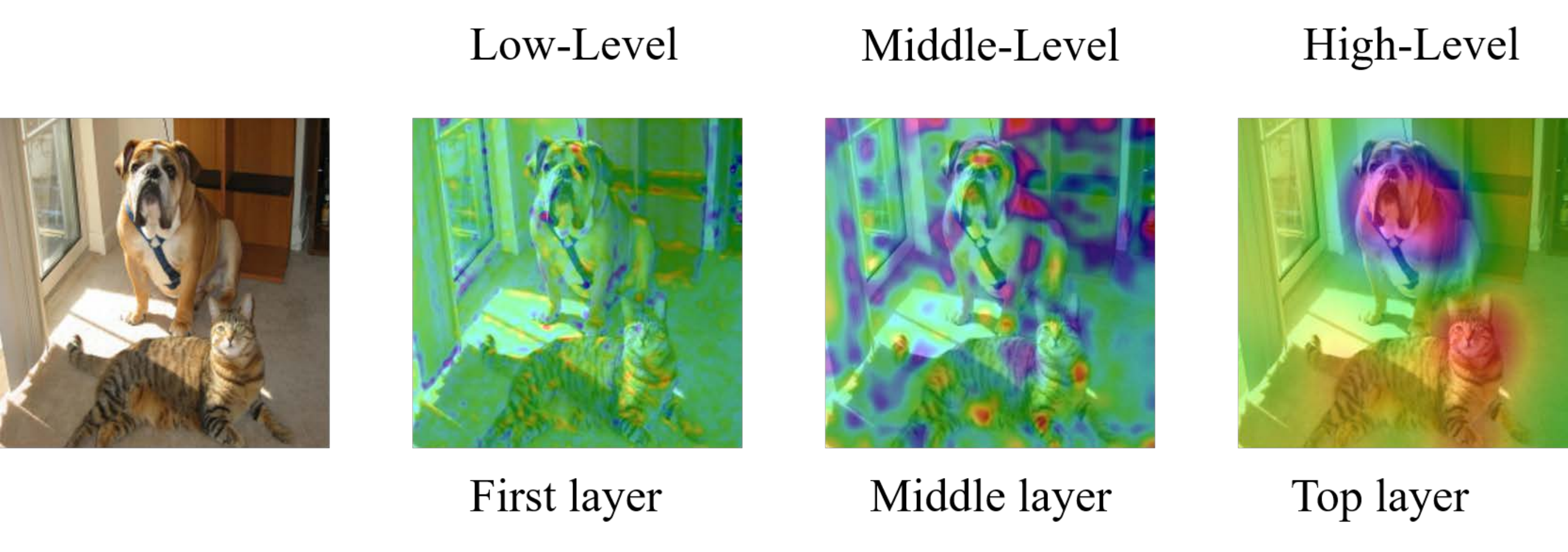}
			\caption{Illustration of the spatial attention map of a streamlined convolutional network.}
			\label{fig:spatial attention map}
		\end{figure}
		
		\subsubsection{Distilling Activation-based Attention}	
		Suppose the importance of a hidden neuron with respect to the particular input is indicated by its absolute value, then we define a function $\mathcal{F}$ that computes statistics of these absolute values across channels to represent the attention knowledge (shown in Fig \ref{fig:spatial attention map}).
		Specifically, we define A $\in$ $R^{C \times H \times M}$ representing the ANN convolutional layer's activation tensor containing $C$ convolutional channels with spatial dimensions of $H \times M$. 
		Then, we use the function $\mathcal{F}$ to map the above $3D$ input tensor into a real value output. 
		In order to discriminate the differences in attention-based knowledge among different targets more clearly, the absolute values are raised to the power of 2. 
		Moreover, this value is averaged by channels and spatial dimensions to ease the impact of individual extreme values and noise on overall performance. 
		The mapping function $\mathcal{F}$ is defined as follows:  
		\begin{equation}
			\label{equation:map_general}
			\mathcal{F}: R^{C \times H \times M} \to R^{H \times M}
		\end{equation}
		\begin{equation}
			\label{equation:map_detail}
			\mathcal{F}_{mean}(A) =\frac{\sum_{i=1}^{C}\sum_{j=1}^{H}\sum_{k=1}^{M} | A_{i,j,k} | ^{2} }{C\times H\times M}
		\end{equation}
		where operations of power and absolute value are element-wise. 
		After introducing attention information for the activation tensor of teacher and student models, we have a new attention loss term $\mathcal{L}_{at}(Ta, St)$: 
		\begin{equation}
			\label{loss_at_act}
			\mathcal{L}_{at}(Ta, St) = \sum_{l\in Z} ||\mathcal{F}^{l}_{mean}(Ta) - \mathcal{F}^{l}_{mean}(St)||_{2}
		\end{equation}
		which indicates the losses of all attention map pairs ($Z$) for both teacher ($Ta$) and student ($St$) networks during distillation process. Therefore, according to Eq. (\ref{loss_at_act}), we have total loss function $\mathcal{L}_{total}$: 
		\begin{equation}
			\label{loss_total_act}
			\mathcal{L}_{total} = \mathcal{L}_{ce} + \frac{\alpha}{2} \mathcal{L}_{at} (Ta, St)
		\end{equation}
		\begin{equation}
			\label{loss_CE}
			\mathcal{L}_{ce} = - \sum_{i} y_{i} log (p_{i})
		\end{equation}
		where $\alpha$ is the hyperparameter, and $\mathcal{L}_{ce}$ denotes the cross-entropy loss between the true output $y$ and the predicted distribution $p$. 
		
		\subsubsection{Distilling Gradient-based Attention}	
		Suppose a pixel's small change results in a significant impact on the model's prediction. The model gives particular attention to that pixel. 
		Then, we define a gradient-based attention, which is viewed as the input sensitivity knowledge learned by the network. 
		In other words, the attention to the input's specific spatial location encodes the sensitivity of the model's output prediction with respect to changes at that location. Thus, the teacher model's gradient of loss in regard to input is defined as follows:
		\begin{equation}
			G_{Ta} = \frac{\partial}{\partial x} \mathcal{L} (W_{Ta}, x)
		\end{equation}
		Inspired by the spike activation map (SAM) \cite{kim2021visual}, we use the spike activity in the forward propagation to define the input sensitivity function. 
		\begin{equation}
			G_{St} (h,m,t) = \sum_{C} \sum_{t' \in O_{h,m}} e^{- |t - t'|} o^{t}_{h,m}
		\end{equation}
		where $t'$ represents the previous spike time, the set $O_{h,m}$ includes previous firing times of a neuron located at $(h, m)$. We then define the total loss as: 
		\begin{equation}
			\label{loss_at_grd}
			\mathcal{L}_{at} (Ta, St) = \sum_{l \in Z} ||G^{l}_{Ta} - G^{l}_{St}||_{2}
		\end{equation}
		\begin{equation}
			\label{loss_total_grd}
			\mathcal{L}_{total} = \mathcal{L}_{ce} + \frac{\alpha}{2} \mathcal{L}_{at} (Ta, St)
		\end{equation}	
		
		Supposing transfer losses are within teacher and student feature maps of the same spatial resolution. 
		However, if necessary, we can use interpolation or downsampling to match their shapes.
	
	\subsection{Layer-wise Supervision Strategy}
		Previous works have shown that various parts of each layer in the model contain different attention information \cite{Parkhi15}. 
		As shown in Fig. \ref{fig:spatial attention map}, with the low layers, neurons reflect intense activation; with the middle layers, neurons' activation tends to focus on recognizable areas, such as feet or eyes; with the top layers, neurons' activation reflects the entire object. 
		
		Recently, ANN-to-SNN knowledge distillation studies mainly focus on minimizing the output distribution between a large ANN and a small SNN \cite{TakuyaZN21, LeeKDSNN}, which fails to achieve satisfactory performance because of lacking enough information for distillation. Generally, when signals pass through every convolution layer, a streamlined convolutional network can output various attention maps containing different levels of attention information. 
		In order to achieve a minimal loss, we divide the learned attention knowledge into three levels, from low to high, to distill attention information sufficiently from a teacher ANN and supervise the transfer losses with the designed loss function (equation \ref{loss_at_act} or \ref{loss_at_grd}). 

	\subsection{Three-stage Training Process of LaSNN}
		The three-stage LaSNN training pipeline is illustrated in Fig. \ref{fig:framework} (b). Firstly, train a teacher (cumbersome) ANN with a bias term, and batch normalization \cite{ioffe2015batch}. 
		
		Secondly, convert a single smaller ANN (intermediate ANN) as the student SNN with IF neurons (the architecture is described in the subsection network architectures). 
		In conversion, we use threshold balancing, where the weights are unchanged and then normalized by the maximum preactivation \cite{Sengupta2019}. 
		
		Thirdly, after finishing the training of the teacher model and conversion of the student model, a layer-wise scheme is adopted for transferring the knowledge of the teacher ANN to the student SNN, the distillation strategy, and loss function designing as described in the above subsections. 
		And the student SNN is optimized with error back-propagation by utilizing a linear surrogate gradient to approximate the discontinuous gradient \cite{bellec2018long}. 
		The pseudo-derivative is described as:  
		\begin{equation}
			\frac{\partial o^{t}_{i}}{\partial v^{t}_{i}} = max\left\{0, 1-|\theta|\right\}
		\end{equation}
		In order to achieve stable performance for deep SNNs, we introduce the decayed term $\gamma<1$ (typically $\gamma=0.3$) that dampens the increase of back-propagated errors through spikes:
		\begin{equation}
			\label{surrogate_grad}
			\frac{\partial o^{t}_{i}}{\partial v^{t}_{i}} = \gamma ~ max\left\{0, 1-|\theta|\right\}
		\end{equation}
		Note that the gradients propagation is not affected, thus that can propagate through many time steps in the dynamic threshold. 
		The weight update is calculated as:
		\begin{equation}
			\Delta \omega_{ij} = \sum_{t} \frac{\partial \mathcal{L}_{total}}{\partial \omega_{ij}} = \sum_{t} \frac{\partial \mathcal{L}_{ce}}{\partial \omega_{ij}} + \sum_{t} \frac{\partial \mathcal{L}_{at}}{\partial \omega_{ij}}
		\end{equation}
		\begin{equation}	
			 \mathcal{L}_{ce} = \sum_{t} \frac{\partial L_{ce}}{\partial o^{t}_{i}} \frac{\partial o^{t}_{i}}{\partial v^{t}_{i}} \frac{\partial v^{t}_{i}}{\partial \omega_{ij}}
		\end{equation}
		where $\partial o^{t}_{i} / {\partial v^{t}_{i}}$ is a non-differentiable term, and we approximate it with the linear surrogate gradient (Equation \ref{surrogate_grad}). The main training process of the student SNN is described in algorithm \ref{alg:training}.
		
		\begin{algorithm}[tb]
			\caption{Overall training algorithm}
			\label{alg:training}
			\textbf{Input}: 
			Pretrained $ANN$ (teacher model, $Ta$), 
			$SNN$ (student model, $St$), 
			input $(X)$, 
			target $(Y)$, 
			weights $(W)$,
			threshold voltage $(V)$,
		    layers $(L)$
			\begin{algorithmic}[1] 
				\STATE \# \ Forward propagation
				\FOR{$t=1$ to $T$}
				\STATE $O_{0}^{t}$ $\leftarrow$ $Poisson~Encoder(X)$ 
				\FOR{$l=1$ to $L-1$}
				\IF{$isinstance (St_{l}, [Conv, Linear])$}
				\STATE \# output sum of the previous layer
				\STATE $V_{l}^{t}=\lambda V_{l}^{t-1} + W_{l}O_{l-1}^{t}-V_{l}^{th} \times O_{l}^{t-1}$
				\STATE \# Generate spike when $V > V_{th}$
				\STATE $O_{l}^{t}$ $\leftarrow$ $Surrogate ~ Gradient(V_{l}^{t}, V_{l}^{th}, t)$
				\IF{$O_{l}^{t} == 1$}
				\STATE \# Save spike times (StT)
				\STATE $StT_{l}^{t} = t$
				\ENDIF
				\ELSIF{$isinstance (St_{l}, AvgPool)$}
				\STATE $O_{l}^{t} = St_{l}(O_{l-1}^{t})$
				\ELSIF{$isinstance (St_{l}, Dropout)$}
				\STATE $O_{l}^{t} = Dropout * O_{l-1}^{t}$
				\ENDIF						
				\ENDFOR
				\STATE $V_{L}^{t}=\lambda V_{L}^{t-1} + W_{L}O_{L-1}^{t}$
				\ENDFOR
				\STATE \# Backward Propagation: compute $\frac{\mathrm{d} \mathcal{L}}{\mathrm{d} V_{L}}$ by Surrogate ~ Gradient
				\FOR{$t=T$ to $1$}
				\FOR{$l=L-1$ to $1$}
				\STATE $\frac{\mathrm{d} \mathcal{L}}{\mathrm{d} V_{l}^{t}} = \frac{\mathrm{d} \mathcal{L}}{\mathrm{d} O_{l}^{t}} \frac{\mathrm{d} O_{l}^{t}}{\mathrm{d} V_{l}^{t}} = \frac{\mathrm{d} \mathcal{L}}{\mathrm{d} O_{l}^{t}} \times \gamma\ max\left\{0, 1-|V_{l}^{th}|\right\} $
				\ENDFOR
				\ENDFOR				
			\end{algorithmic}
		\end{algorithm}

\section{Experiments}
	\subsection{Data sets and Settings}	
		We evaluate the performance of the LaSNN framework on CIFAR-10, CIFAR-100, and Tiny ImageNet data sets.		
		\begin{itemize}
			\item
			\textbf{CIFAR-10}: The data set contains $60,000$ labeled images of $10$ categories, divided into training $(50,000)$ and testing $(10,000)$ sets. The images size is $32\times32$, and has RGB three channels.
			\item
			\textbf{CIFAR-100}: The data set is similar to CIFAR-10 except that it contains $100$ categories.
			\item
			\textbf{Tiny ImageNet}: The data set is the subset of ImageNet containing $200$ categories. Each class has $500$ training images, $50$ validation images and $50$ testing images. The images are of size $64\times64$ with RGB channels. 
		\end{itemize}
		
		Both the teacher ANN and the intermediate ANN used for converting to the student SNN are trained from scratch, setting batch size as $256$ and employing SGD optimizer. The learning rate is $0.01$, and the weight decay is $0.0005$. The teacher ANN reaches convergence after $300$ epochs, while the intermediate ANN reaches convergence after $200$ epochs. The time step is set to $2500$ during the conversion process. After converting, use the LaSNN framework to train the student SNN. 
		We use the Adam optimizer \cite{kingma2014adam} in the process of training the student SNN with the layer-wise distillation (proposed work) and label-based distillation \cite{TakuyaZN21}, setting the learning rate as $0.0001$ and the weight decay as $0.0005$. The decayed term $\alpha$ for attention losses is set to $0.9$, according to Hinton et al. \cite{Hinton2015}. For the student SNN, setting training epoch as $100$, the mini-batch size as $16$, and the time step as $100$. All performances are evaluated on NVIDIA GeForce RTX 3090 GPU that has 24 GB of memory. 
		
		\begin{table*}[ht]
			\normalsize
			\renewcommand{\arraystretch}{1.3}
			\centering
			\caption{Comparison of LaSNN and Other Converted SNNs.}
			\label{tab:comparison}
			\begin{tabular}{ccccc}
				\hline
				Model                          & Training Method     & Architecure       & Time Steps   & Accuracy         \\ \hline
				\multicolumn{5}{c}{CIFAR-10} \\ \hline
				Hunsverger \& Eliasmith (2015) & ANN-to-SNN Conversion  & 2Conv, 2Linear    & 6000         & 82.95\%          \\ 
				Cao et al. (2015)              & ANN-to-SNN Conversion  & 3Conv, 2Linear    & 400          & 77.43\%          \\ 
				Sengupta et al. (2019)         & ANN-to-SNN Conversion  & VGG16             & 2500         & 91.55\%          \\ 
				\textbf{This work}             & \textbf{LaSNN} & \textbf{3Conv, 2Linear} & \textbf{100} & \textbf{83.27\%} \\  
				\textbf{This work}             & \textbf{LaSNN} & \textbf{VGG16} & \textbf{100} & \textbf{91.22\%} \\
				\hline
		\end{tabular}
		\end{table*}	
	
	\subsection{Performance and Comparison}
		We evaluate the proposed LaSNN framework with activation-based and gradient-based distillation on CIFAR-10, CIFAR-100, and TinyImageNet data sets. All the teacher models have widely used ANN architectures (DenseNet121, VGG16, and ResNet20), and the student models are relatively shallower ANN-like architectures. For simplicity, we first employ activation-based distillation for the comparison experiments. And then, we explore the effects of activation-based and gradient-based distillation.

		\subsubsection{Comparison with Non-distillation Algorithms}		
			In order to demonstrate the effectiveness of our LaSNN framework, LaSNN is compared to three main non-distillation SNN training algorithms under the same conditions on the CIFAR-10 data set, including the ANN-to-SNN conversion algorithm, hybrid training algorithm, and calibration algorithm. 
			
			As shown in Table \ref{tab:comparison}, on the CIFAR-10 data set, LaSNN achieves higher performance than ANN-to-SNN conversion models with relatively shallow structures and comparable performance to converted SNNs with deep architectures. 
			Notably, LaSNN is significantly more efficient than ANN-to-SNN conversion methods both in shallow and deep architectures (e.g., VGG16). The advanced accuracy and efficiency performance of LaSNN are attributed to its layer-wise distillation process with the attention scheme to represent information both in teacher ANNs and student SNNs.
			
			Then, we further investigate the effectiveness of our LaSNN framework by carefully comparing LaSNN with the hybrid training algorithm \cite{RathiSP020}. 
			Detailed results are summarized in Table \ref{tab:general}, LaSNN is significantly better than hybrid-trained SNNs not only in shallow structures but also in deep architectures over different data sets (CIFAR-10, CIFAR-100 and Tiny ImageNet). 
			In other words, by forcing the student SNN to imitate the attention maps of a cumbersome (teacher) ANN, the performance of a student SNN can be significantly improved. Notably, LaSNN achieves more improved performance when tasks become more complex (e.g., on the Tiny ImageNet data set), illustrating that our LaSNN framework may be promising for more challenging tasks.
			
			\begin{figure}[h]
				\centering
				\includegraphics[width=0.5\textwidth]{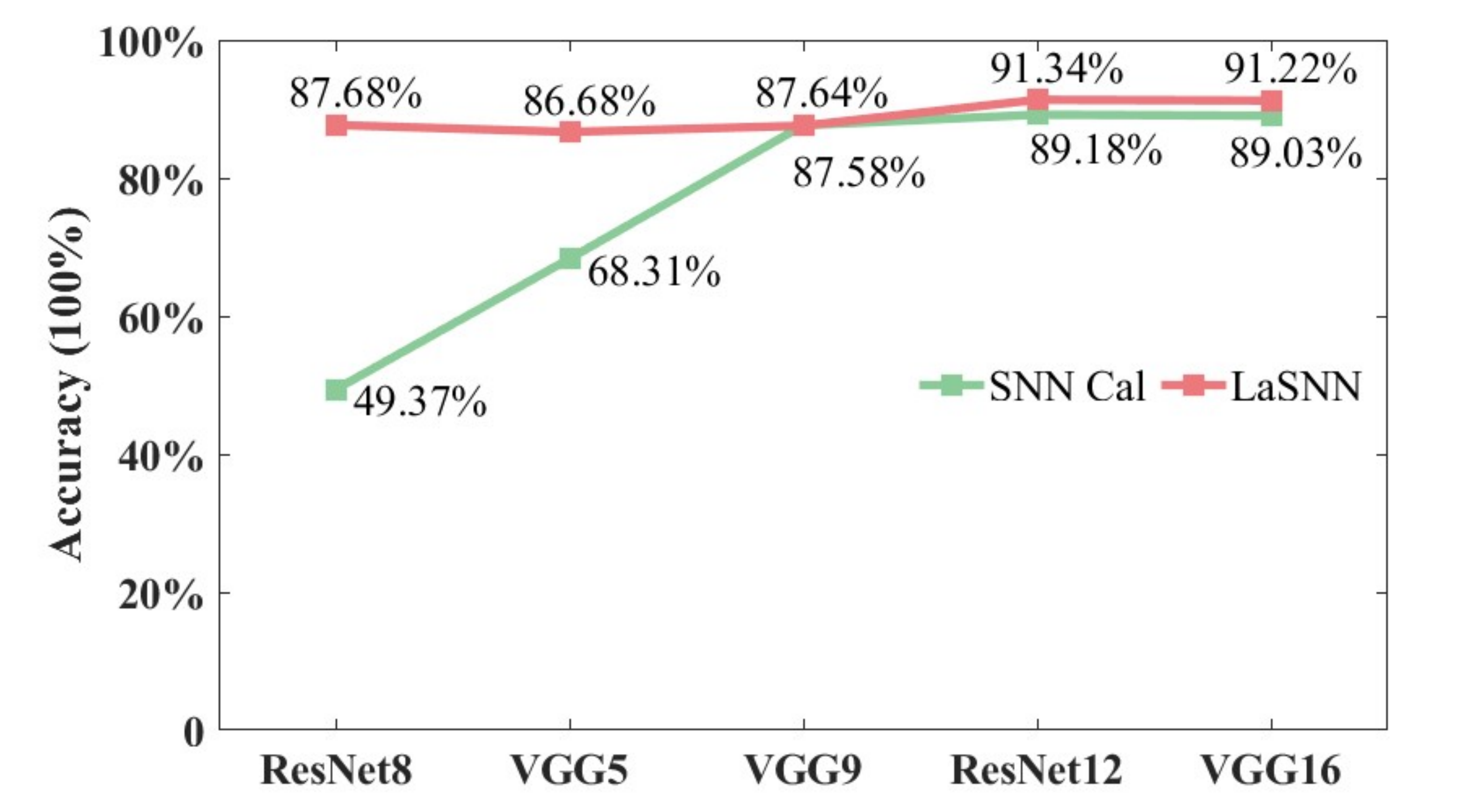}
				\caption{Performance comparison of the LaSNN framework and the SNN calibration algorithm on the CIFAR-10 data set.}
				\label{fig:snn-cal comparison}
			\end{figure}
		
			\begin{table*}[h]
				\normalsize
				\renewcommand{\arraystretch}{1.3}
				\centering
				\caption{Top-1 Classification Accuracy.}
				\label{tab:general}
					\begin{tabular}{ccccc}
						\hline
						SNN Architecture & ANN & Hybrid training \cite{RathiSP020}& ANN-to-SNN distillation \cite{TakuyaZN21} & \textbf{LaSNN} \\ \hline
						\multicolumn{5}{c}{CIFAR-10} 						\\ \hline
						3Conv, 2Linear & 80.38\% & 82.05\% & 82.27\% & \textbf{83.65\%} \\ 
						VGG5 & 87.58\% & 85.76\% & 85.94\% & \textbf{86.68\%} \\
						VGG9     & 88.84\% & 86.35\% & 86.39\% & \textbf{87.58\%} \\
						ResNet12 & 92.46\% & 90.70\% & 90.72\% & \textbf{91.34\%} \\ 
						VGG16 & 92.35\% & 90.53\% & 90.60\% & \textbf{91.22\%} \\ \hline
						\multicolumn{5}{c}{CIFAR-100}  					\\ \hline
						VGG7     & 64.61\% & 60.69\% & 60.83\% & \textbf{61.52\%} \\ \hline
						\multicolumn{5}{c}{Tiny ImageNet}					\\ \hline 
						VGG16    & 48.39\% & 39.48\% & 45.49\% & \textbf{47.44\%} \\ \hline
					\end{tabular}
				\end{table*} 
			
			In addition, we conduct the performance comparison between LaSNN and the calibration algorithm on various networks. 
			Details are shown in Fig. \ref{fig:snn-cal comparison}, the inference accuracy of SNNs decreases as the network structures become shallow, but LaSNN is significantly more stable than the calibration algorithm of classification accuracy for all networks. 
			In contrast, the performance of the calibration algorithm decreases obviously for small architectures, illustrating that our approach is more scalable to SNNs than the calibration algorithm. 
	
		\subsubsection{Comparison with Distillation Algorithms}		
			In this case, we compare our approach with two distillation algorithms (label-based distillation \cite{TakuyaZN21} and spike-based distillation \cite{KushawahaKBV20}) to evaluate the efficacy of our LaSNN framework. 
			
			Compared with LaSNN, ANN-to-SNN distillation only distills label-based knowledge of the last layer from pre-trained ANNs to SNNs. The results are shown in Table \ref{tab:general}. On all three data sets, LaSNN shows significant improvements and reaches comparable performance to ANNs corresponding improvements are 0.62\% to 1.38\% on the CIFAR-10, 0.69\% on the CIFAR-100, and 1.95\% on the Tiny ImageNet data set. 
		
			The spike-based knowledge distillation algorithm constructs a three-dimensional matrix as the spiking activation tensor that is used to represent and transfer the knowledge from the outputs of the teacher SNN to the student SNN \cite{KushawahaKBV20}. 
			As shown in Table \ref{tab:comparison_spike}, LaSNN shows remarkable improvements in inference performance compared to distilling spikes with similar small architectures on the CIFAR-10 data set. Along with the spiking activation tensor, the spiking distilled method uses the sliding window to accumulate the losses over the total time steps ($T$) during the distillation process, consuming a lot of computational resources. Moreover, the authors in Kushawaha et al. \cite{KushawahaKBV20}, employed a multi-stage distillation scheme \cite{mirzadeh2020improved} to improve the classification accuracy of student SNNs, which introduces resource consumption further. However, our LaSNN framework employed a hybrid training scheme that overcomes the inherent low-performance and high-latency problems of deep SNNs. In addition, attention-based knowledge contains much more helpful information than one-hot labels. 
			\begin{table}[h]
				\normalsize
				\renewcommand{\arraystretch}{1.3}
				\centering
				\caption{Comparison of LaSNN and the spike-based knowledge distillation (KD) algorithm.}
				\label{tab:comparison_spike}
				\begin{threeparttable}
					\begin{tabular}{ccc}
						\hline
						Model                     & Architecture & Accuracy \\ \hline
						\multicolumn{3}{c}{CIFAR-10} \\ \hline
						Spike-based KD \cite{KushawahaKBV20} & Ta-Ass-St    & 42.38\%  \\ 
						\textbf{This work}        & \textbf{3Conv, 2Linear} & \textbf{83.65\%}  \\ 
						\textbf{This work}        & \textbf{VGG5}           & \textbf{86.68\%}  \\ 
						\textbf{This work}        & \textbf{VGG9}           & \textbf{87.58\%}  \\
						\hline
					\end{tabular}
				\begin{tablenotes} 
					\item Note: teacher network(Ta), teacher assistant network(Ass), student network(St).
				\end{tablenotes}
			\end{threeparttable}
			\end{table}	
			
		The above results emphasize the validity of distilling attention information from ANNs to SNNs with the layer-wise scheme, which is essential for improving the performance of SNNs.

	\subsection{Effect of Distillation Strategies and Different Teacher ANNs}
		In this part, ablation experiments are conducted to demonstrate the importance of layer-wise distillation in recognition performance. 
		Then, we evaluate the effectiveness of two attention-based distillation strategies, i.e., activation-based attention and gradient-based attention. 
		Finally, we evaluate the impact of using different teacher ANNs.
		
		\subsubsection{Effectiveness of Layer-wise Distillation}
			In order to show the importance of layer-wise distillation scheme, we first conduct an ablation experiment with various network structures (ResNet12, VGG7, and VGG16) on the CIFAR-10, CIFAR-100, and Tiny ImageNet data sets. Fig \ref{fig:framework} (b) describes two processes employed in the training pipeline of student SNNs, initialize an SNN model from an intermediate ANN by the parameter normalization method, then optimize the initialized SNN through a layer-wise distillation scheme. In our experiment, we compare LaSNN with two models: 1) the model without LaSNN: the initial conversion SNN; and 2) the model without layer-wise distillation scheme: directly trained the initial conversion SNN using the same surrogate gradient. For simplicity, we employ the activation-based distillation scheme in LaSNN, similar to the above comparison experiments.
			
\begin{figure}[h]
				\centering
				\includegraphics[width=0.5\textwidth]{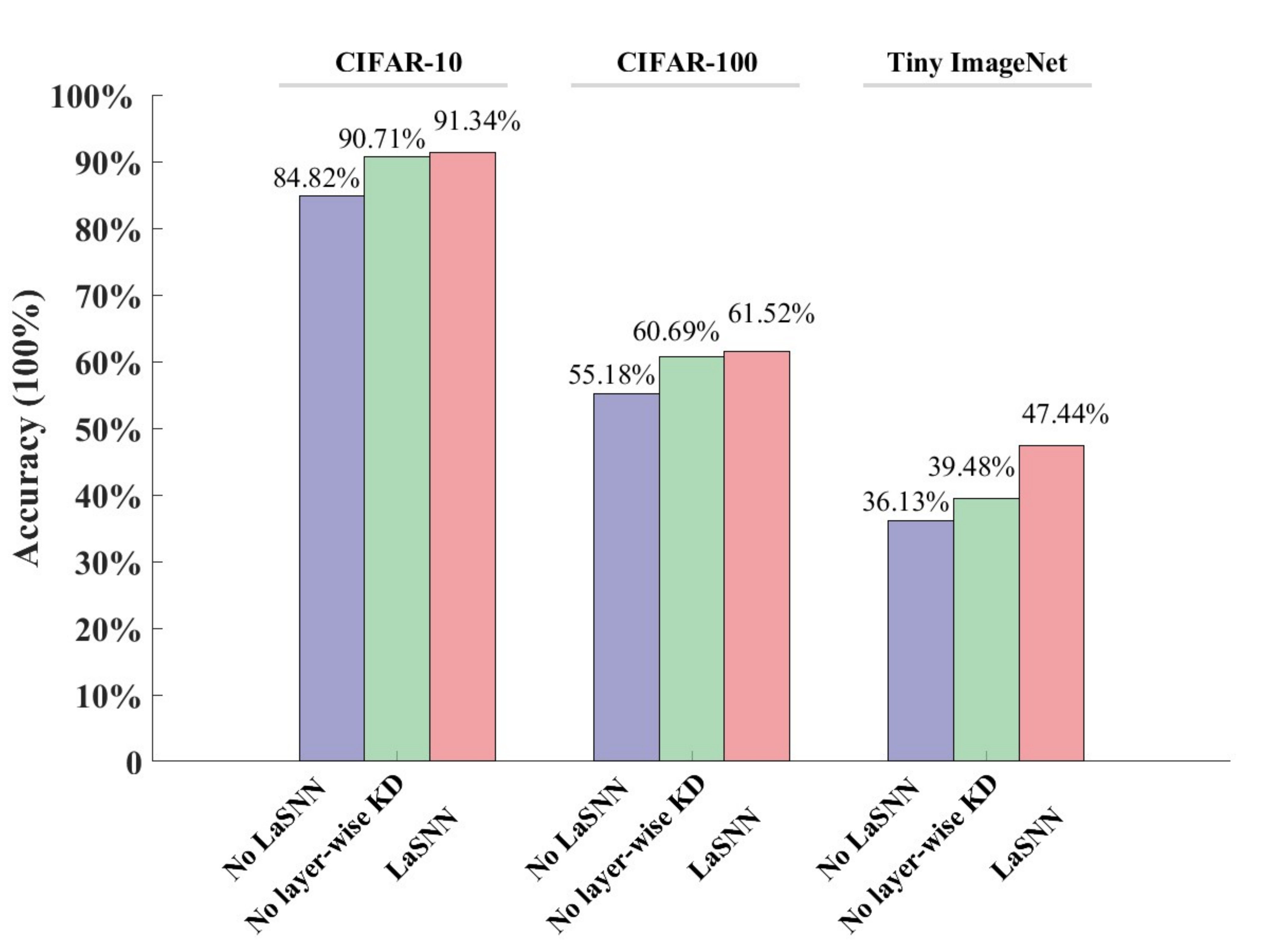}
				\caption{Ablation experiments of the LaSNN framework on different data sets. }
				\label{fig:ablation}
			\end{figure}

			Fig. \ref{fig:ablation} shows the effects of the layer-wise distillation scheme on the accuracy performance. When non-LaSNN training scheme is employed, the training algorithm is degraded into an ANN-to-SNN conversion algorithm. 
			In the meantime, when the non-layer-wise scheme is employed, the training algorithm is degraded into the hybrid training method, resulting in performance decreased in all three networks over three data sets. 
			
			Notably, as shown in Fig. \ref{fig:ablation}, if employing the layer-wise scheme, the accuracy performance increases significantly on the Tiny ImageNet task. This indicates layer-wise distillation scheme is promising for more challenging tasks.  		
		
		\subsubsection{Effectiveness of Two Attention-based Distillation Strategies}
\begin{figure}[h]
				\centering
				\includegraphics[width=0.5\textwidth]{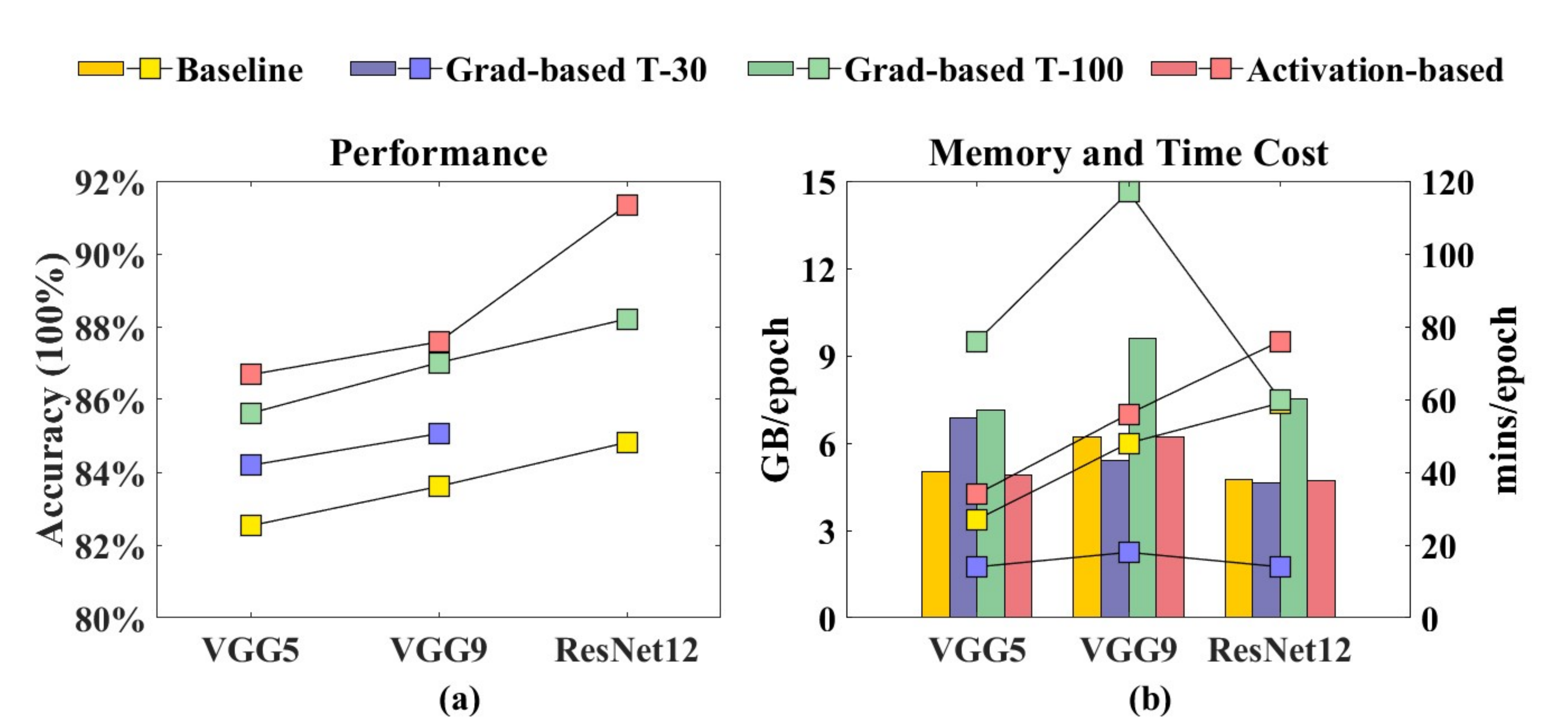}
				\caption{Performance, training time, and memory cost comparison of two attention-based distillation schemes on the CIFAR-10 data set. (a) Comparison of performance. (b) Comparison of training time (curve) and memory cost (bar graph) for two attention-based schemes. The abbreviation Grad-based T-30 (T-100): stands for using the gradient-based attention method with 30 (100) time steps.}
				\label{fig:act_grad}
			\end{figure}

			To check whether distilling knowledge from activation-based attention can be more beneficial than from gradient-based attention, we train three networks with different depths (VGG5, VGG9, ResNet12) on the CIFAR-10 data set, and two attention-based distillation schemes are employed. 
			Deterministic algorithms and the fixed random seed are adopted in the experiments. 
			Both activation-based and gradient-based schemes are evaluated under the same experimental settings (details are provided in the experiments section). Accuracy results are shown in Fig.\ref{fig:act_grad} (a). Similar to activation-based attention, employing the gradient-based attention scheme leads to improved performance. However, compared with the activation-based scheme in the same training settings, the gradient-based scheme achieves weaker performance. 
			
			In addition, as the gradient-based scheme needs to calculate gradient maps of the teacher ANN and spiking activation maps of the student SNN, LaSNN with the gradient-based scheme consumes more computing and memory resources, shown in Fig.\ref{fig:act_grad} (b). More specifically, one epoch of gradient-based (activation-based) training with VGG5$/$VGG9$/$ResNet12 structures takes $76/117/60$ ($34/56/76$) minutes and $7.13/9.59/7.51$ ($4.89/6.23/4.71$) GB of GPU memory, respectively. 
			
			We find that the most resource-intensive operation is calculating the SAM, which needs to compute the activation maps per time step and accumulate the values to evaluate the contribution of previous spikes with respect to the current neural state. Thus, we additionally trained the same student SNNs (VGG5$/$VGG9$/$ResNet12) with 30 time steps. Although they achieve weaker performance than same distillation models with 100 time steps, the memory requirements are reduced to the same level as activation-based scheme, shown in Fig.\ref{fig:act_grad} (b). In the future, we plan to explore more efficient gradient-based attention for ANN-to-SNN distillation because it is so far unclear how gradient-based attention transfer in the form of spikes. 
		
		\subsubsection{Performance Using Different Teacher ANNs}
			It was reported that knowledge distillation suffers severe performance losses when the structures and depths of the teacher and student networks are different \cite{ZagoruykoK17}. 
			Thus, we investigate the performances of LaSNN with various $ANN/SNN$ pairs based on the CIFAR-10 data set. The detailed results of all combinations are given in Table \ref{tab:diff-teacher}. We chose three widely used ANNs (DenseNet121, VGG16, and ResNet20) as the teacher ANN. The VGG5 and VGG9 student SNN achieve the same performance for the teacher ANN with different structures (DenseNet121 and ResNet20 teacher ANN) and different depths (VGG16 teacher ANN). Although the ResNet12 student SNN achieves different performances with different teacher ANNs, the accuracy effectively increased compared to the non-LaSNN training scheme (84.82\%) and reached comparable performance to ANN (92.46\%). The above results highlight the potential that LaSNN can effectively transfer knowledge from various ANNs to SNNs. More importantly, LaSNN is flexible for training deep SNNs. Thus, it is not necessary to train corresponding assistant networks during distillation process \cite{ZagoruykoK17, KushawahaKBV20}. 		
			\begin{table*}[h]
				\normalsize
				\renewcommand{\arraystretch}{1.3}
				\centering
				\caption{Classification results (Top-1) with different teacher ANN/student SNN pairs on the CIFAR-10 data set.}
				\label{tab:diff-teacher}
				\begin{tabular}{cccccc}
					\hline
					SNN Architecture & ANN & Without LaSNN & \multicolumn{3}{c}{LaSNN with Different Teacher ANNs} \\ \hline
					&         &         & DenseNet121 (95.84\%) & VGG16 (91.53\%) & ResNet20 (93.04\%) \\ 
					VGG5 & 87.58\% & 82.53\% & 86.66\% & 86.68\% & 86.61\% \\
					VGG9 & 88.84\% & 83.61\% & 87.57\% & 87.58\% & 87.55\% \\
					ResNet12 & 92.46\% & 84.82\% & 91.34\% & 90.90\% & 91.23\% \\ 			    \hline
				\end{tabular}
			\end{table*}

	\subsection{Analysis of Input Encoding and Computational Cost}
		The impact of input encoding and the comparison of computing efficiency are analyzed in this section. 
		In this section, we focus on investigating the impact of input encoding and comparing the computing efficiency of the LaSNN framework and other SNN training methods. 
	
		\subsubsection{Analysis of Input Encoding}
			In rate coding, such as Poisson encoding, the analog values indicate neurons' firing rate. In the meantime, the neuron outputs binary values (0 or 1) at each time step in SNNs. Specifically, an analog value of 0.3 indicates a neuron firing spikes during 30\% total time steps. 
			Thus, obtaining high accuracy requires adopting an enormous number of time steps, which leads to high inference latency. 
			Recent studies in state-of-the-art SNNs employ the direct encoding scheme to reduce inference latency \cite{wu2019direct, rathi2021diet}. 
			
\begin{figure}[h]
				\centering
				\includegraphics[width=0.5\textwidth]{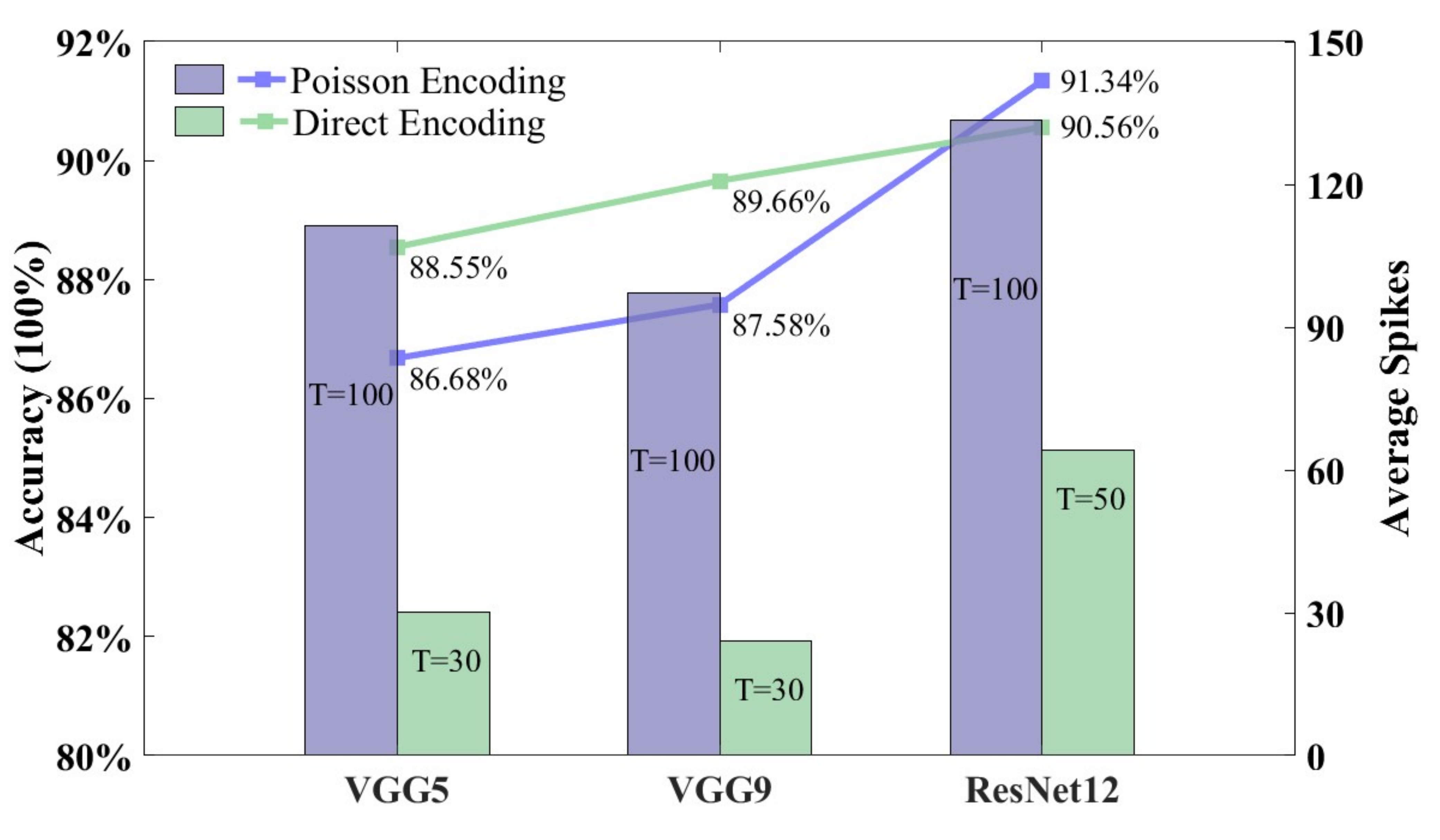}
				\caption{Effects of employing direct input encoding and Poisson input encoding on the CIFAR-10 data set. The abbreviation T: stands for time steps.}
				\label{fig:direct_exp}
			\end{figure}

			In this part, we analyze the impact of two different input encoding methods on the average spike rate. 
			We train two different student SNNs with VGG5, VGG9, and ResNet12 structures and two input encoding methods: 1) Poisson rate encoding; 2) direct input encoding. 
			Firstly, the SNNs are trained under the same conditions (such as teacher-student pairs, learning rate, and the number of iterations) for evaluating performance. 
			Next, the same student SNNs with different input encoding schemes learned from the same teacher ANN are trained to achieve similar accuracy on the CIFAR-10 data set for evaluating the average spike rate. 
			As shown in Fig.\ref{fig:direct_exp}, the performance of VGG5 and VGG9 student SNN with direct input encoding are higher than corresponding SNNs with Poisson input encoding ($88.55\%/89.66\%$ vs. $86.68\%/87.58\%$). 
			Although ResNet12 student SNN with direct input encoding achieves lower performance than Poisson input encoding ($90.56\%$ vs. $91.34\%$, shown in Fig.\ref{fig:direct_exp}), it can reach a level significantly close to using hybrid training and ANN-to-SNN distillation methods ($90.70\%$ and $90.72\%$, in Table \ref{tab:general}). 
			The results indicate that LaSNN is feasible for direct input encoding. 
			
			In addition, the student SNNs with Poisson input encoding and 100 time steps achieve average spikes of $111.25/97.16/133.37$ (after evaluating 2000 samples from the CIFAR-10 test set, sum the spike in all time steps, and then divide the total number of neurons to calculate the average number of spikes). 
			As shown in Fig.\ref{fig:direct_exp}, when we replace Poisson input encoding with direct input encoding, the inference latency is reduced to $30/50$ time steps. 
			With Poisson input encoding, the firing rate is proportional to the input pixel value. In contrast, with direct input encoding, the pixel value is fed into the first layer as the input current, and the spike is generated through the IF or LIF spiking neuron. 
			Therefore, direct input encoding reduces the time steps required to encode the input. 
				
		\subsubsection{Analysis of Computational Cost}
\begin{figure}[h]
			\centering
			\includegraphics[width=0.5\textwidth]{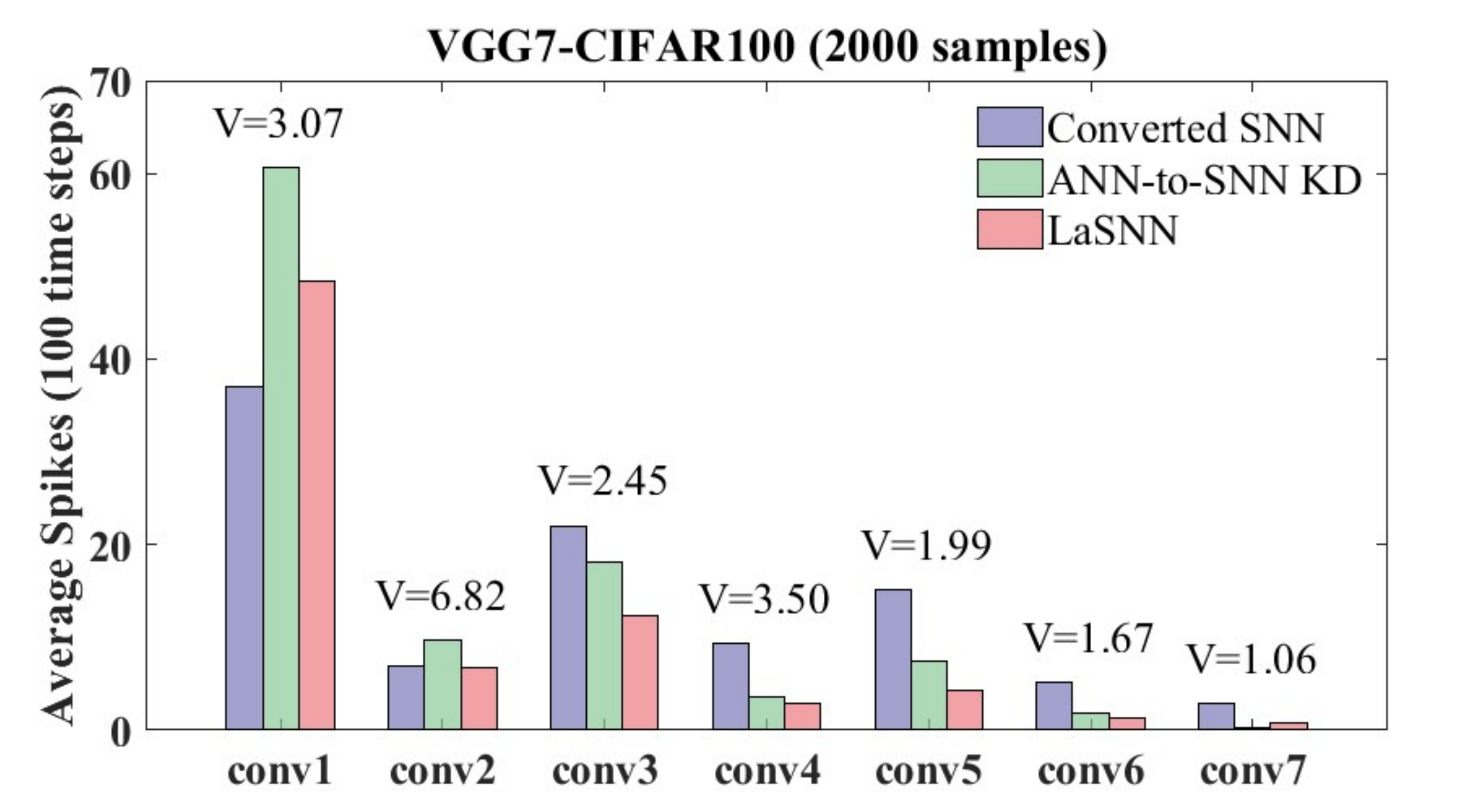}
			\caption{Energy analysis on the CIFAR-100 data set over 2000 samples. The abbreviation V: stands for voltage threshold.}
			\label{fig:efficiency}
		\end{figure}

			Since the energy consumption of a single spike in an SNN is constant \cite{CaoCK15}, the fundamental energy consumption analysis depends on the number of spikes as well as the total number of time steps. 
			The result in Fig. \ref{fig:efficiency} illustrates the average number of spikes in every convolutional layer of SNNs with the VGG7 architecture after evaluating 2000 samples from the CIFAR-100 test set. 
			The average number of spikes is compared on a converted SNN, a distillation SNN, and the LaSNN framework, the more intense the spiking activity is, the more energy is consumed. 
			Under the same conditions, including inputs, time steps, threshold voltages, and others, the LaSNN framework achieves fewer average spikes in most layers and obtains higher performance compared to the converted SNN and the distillation SNN. 

\section{Conclusion}
	In this work, we first abstracted an attention representation to bridge the information gap between ANNs and SNNs during the ANN-to-SNN distillation process. 
	Then, we extended the ANN-to-SNN distillation with our layer-wise scheme. 
	Moreover, three processes of training were introduced to optimize the weights of the student SNN. 
	Finally, we propose a new LaSNN framework for ANN-to-SNN distillation utilizing a simple and practical paradigm to effectively and efficiently transfer the learning, in contrast to distilling label-based information from the last layer in other studies. 
	We evaluated the performance of our new framework with various architectures and two different input encoding methods on three benchmark data sets. 
	Experimental results demonstrated that LaSNN achieved competitive top-1 accuracy compared to ANNs and 20x faster inference than converted SNNs with similar performance. Ablation experiments illustrated that our layer-wise scheme plays a crucial role in effectively and efficiently distilling the knowledge. In addition, LaSNN is practical for ANNs and SNNs with different architectures/depths and encoding methods. 
	Accordingly, our LaSNN framework will be more beneficial for developing accurate, efficient, and scalable deep SNNs compared to other SNN training schemes.


\bibliographystyle{IEEEtran}

\bibliography{tnnls}

\begin{thebibliography}{10}
\providecommand{\url}[1]{#1}
\csname url@samestyle\endcsname
\providecommand{\newblock}{\relax}
\providecommand{\bibinfo}[2]{#2}
\providecommand{\BIBentrySTDinterwordspacing}{\spaceskip=0pt\relax}
\providecommand{\BIBentryALTinterwordstretchfactor}{4}
\providecommand{\BIBentryALTinterwordspacing}{\spaceskip=\fontdimen2\font plus
\BIBentryALTinterwordstretchfactor\fontdimen3\font minus
  \fontdimen4\font\relax}
\providecommand{\BIBforeignlanguage}[2]{{%
\expandafter\ifx\csname l@#1\endcsname\relax
\typeout{** WARNING: IEEEtran.bst: No hyphenation pattern has been}%
\typeout{** loaded for the language `#1'. Using the pattern for}%
\typeout{** the default language instead.}%
\else
\language=\csname l@#1\endcsname
\fi
#2}}
\providecommand{\BIBdecl}{\relax}
\BIBdecl

\bibitem{Ghosh-Dastidar2009}
S.~Ghosh-Dastidar and H.~Adeli, ``Third generation neural networks: Spiking
  neural networks,'' in \emph{Advances in Computational Intelligence}, W.~Yu
  and E.~N. Sanchez, Eds.\hskip 1em plus 0.5em minus 0.4em\relax Berlin,
  Heidelberg: Springer Berlin Heidelberg, 2009, pp. 167--178.

\bibitem{TrueNorth}
F.~Akopyan, J.~Sawada, A.~Cassidy, R.~Alvarez-Icaza, J.~Arthur, P.~Merolla,
  N.~Imam, Y.~Nakamura, P.~Datta, G.-J. Nam, B.~Taba, M.~Beakes, B.~Brezzo,
  J.~B. Kuang, R.~Manohar, W.~P. Risk, B.~Jackson, and D.~S. Modha,
  ``Truenorth: Design and tool flow of a 65 mw 1 million neuron programmable
  neurosynaptic chip,'' \emph{IEEE Transactions on Computer-Aided Design of
  Integrated Circuits and Systems}, vol.~34, no.~10, pp. 1537--1557, 2015.

\bibitem{navaridas2015spinnaker}
J.~Navaridas, M.~Luj{\'a}n, L.~A. Plana, S.~Temple, and S.~B. Furber,
  ``Spinnaker: Enhanced multicast routing,'' \emph{Parallel Computing},
  vol.~45, pp. 49--66, 2015.

\bibitem{zou2020hybrid}
Z.~Zou, R.~Zhao, Y.~Wu, Z.~Yang, L.~Tian, S.~Wu, G.~Wang, Y.~Yu, Q.~Zhao,
  M.~Chen \emph{et~al.}, ``A hybrid and scalable brain-inspired robotic
  platform,'' \emph{Scientific reports}, vol.~10, no.~1, pp. 1--13, 2020.

\bibitem{tavanaei2019deep}
A.~Tavanaei, M.~Ghodrati, S.~R. Kheradpisheh, T.~Masquelier, and A.~Maida,
  ``Deep learning in spiking neural networks,'' \emph{Neural networks}, vol.
  111, pp. 47--63, 2019.

\bibitem{diehl2015unsupervised}
P.~U. Diehl and M.~Cook, ``Unsupervised learning of digit recognition using
  spike-timing-dependent plasticity,'' \emph{Frontiers in computational
  neuroscience}, vol.~9, p.~99, 2015.

\bibitem{diehl2015fast}
P.~U. Diehl, D.~Neil, J.~Binas, M.~Cook, S.-C. Liu, and M.~Pfeiffer,
  ``Fast-classifying, high-accuracy spiking deep networks through weight and
  threshold balancing,'' in \emph{2015 International joint conference on neural
  networks (IJCNN)}.\hskip 1em plus 0.5em minus 0.4em\relax ieee, 2015, pp.
  1--8.

\bibitem{kheradpisheh2018stdp}
S.~R. Kheradpisheh, M.~Ganjtabesh, S.~J. Thorpe, and T.~Masquelier,
  ``Stdp-based spiking deep convolutional neural networks for object
  recognition,'' \emph{Neural Networks}, vol.~99, pp. 56--67, 2018.

\bibitem{david1986learning}
E.~R. David, E.~H. Geoffrey, and J.~W. Ronald, ``Learning representations by
  back-propagating errors,'' \emph{nature}, vol. 323, no. 6088, pp. 533--536,
  1986.

\bibitem{wu2019direct}
Y.~Wu, L.~Deng, G.~Li, J.~Zhu, Y.~Xie, and L.~Shi, ``Direct training for
  spiking neural networks: Faster, larger, better,'' in \emph{Proceedings of
  the AAAI Conference on Artificial Intelligence}, vol.~33, no.~01, 2019, pp.
  1311--1318.

\bibitem{NeftciMZ19}
\BIBentryALTinterwordspacing
E.~O. Neftci, H.~Mostafa, and F.~Zenke, ``Surrogate gradient learning in
  spiking neural networks: Bringing the power of gradient-based optimization to
  spiking neural networks,'' \emph{{IEEE} Signal Process. Mag.}, vol.~36,
  no.~6, pp. 51--63, 2019. [Online]. Available:
  \url{https://doi.org/10.1109/MSP.2019.2931595}
\BIBentrySTDinterwordspacing

\bibitem{CaoCK15}
\BIBentryALTinterwordspacing
Y.~Cao, Y.~Chen, and D.~Khosla, ``Spiking deep convolutional neural networks
  for energy-efficient object recognition,'' \emph{Int. J. Comput. Vis.}, vol.
  113, no.~1, pp. 54--66, 2015. [Online]. Available:
  \url{https://doi.org/10.1007/s11263-014-0788-3}
\BIBentrySTDinterwordspacing

\bibitem{Sengupta2019}
\BIBentryALTinterwordspacing
A.~Sengupta, Y.~Ye, R.~Wang, C.~Liu, and K.~Roy, ``Going deeper in spiking
  neural networks: Vgg and residual architectures,'' \emph{Frontiers in
  Neuroscience}, vol.~13, 2019. [Online]. Available:
  \url{https://www.frontiersin.org/articles/10.3389/fnins.2019.00095}
\BIBentrySTDinterwordspacing

\bibitem{han2020rmp}
B.~Han, G.~Srinivasan, and K.~Roy, ``Rmp-snn: Residual membrane potential
  neuron for enabling deeper high-accuracy and low-latency spiking neural
  network,'' in \emph{Proceedings of the IEEE/CVF conference on computer vision
  and pattern recognition}, 2020, pp. 13\,558--13\,567.

\bibitem{deng2021optimal}
S.~Deng and S.~Gu, ``Optimal conversion of conventional artificial neural
  networks to spiking neural networks,'' \emph{arXiv preprint
  arXiv:2103.00476}, 2021.

\bibitem{Hinton2015}
G.~Hinton, O.~Vinyals, and J.~Dean, ``Distilling the knowledge in a neural
  network,'' \emph{Computer Science}, vol.~14, no.~7, pp. 38--39, 2015.

\bibitem{ZagoruykoK17}
\BIBentryALTinterwordspacing
S.~Zagoruyko and N.~Komodakis, ``Paying more attention to attention: Improving
  the performance of convolutional neural networks via attention transfer,'' in
  \emph{5th International Conference on Learning Representations, {ICLR} 2017,
  Toulon, France, April 24-26, 2017, Conference Track Proceedings}.\hskip 1em
  plus 0.5em minus 0.4em\relax OpenReview.net, 2017. [Online]. Available:
  \url{https://openreview.net/forum?id=Sks9\_ajex}
\BIBentrySTDinterwordspacing

\bibitem{taherkhani2018supervised}
A.~Taherkhani, A.~Belatreche, Y.~Li, and L.~P. Maguire, ``A supervised learning
  algorithm for learning precise timing of multiple spikes in multilayer
  spiking neural networks,'' \emph{IEEE transactions on neural networks and
  learning systems}, vol.~29, no.~11, pp. 5394--5407, 2018.

\bibitem{tavanaei2016bio}
A.~Tavanaei and A.~S. Maida, ``Bio-inspired spiking convolutional neural
  network using layer-wise sparse coding and stdp learning,'' \emph{arXiv
  preprint arXiv:1611.03000}, 2016.

\bibitem{amir2017low}
A.~Amir, B.~Taba, D.~Berg, T.~Melano, J.~McKinstry, C.~Di~Nolfo, T.~Nayak,
  A.~Andreopoulos, G.~Garreau, M.~Mendoza \emph{et~al.}, ``A low power, fully
  event-based gesture recognition system,'' in \emph{Proceedings of the IEEE
  conference on computer vision and pattern recognition}, 2017, pp. 7243--7252.

\bibitem{LeeKDSNN}
\BIBentryALTinterwordspacing
D.~Lee, S.~Park, J.~Kim, W.~Doh, and S.~Yoon, ``Energy-efficient knowledge
  distillation for spiking neural networks,'' \emph{CoRR}, vol. abs/2106.07172,
  2021. [Online]. Available: \url{https://arxiv.org/abs/2106.07172}
\BIBentrySTDinterwordspacing

\bibitem{diehl2016truehappiness}
P.~U. Diehl, B.~U. Pedroni, A.~Cassidy, P.~Merolla, E.~Neftci, and G.~Zarrella,
  ``Truehappiness: Neuromorphic emotion recognition on truenorth,'' in
  \emph{2016 international joint conference on neural networks (ijcnn)}.\hskip
  1em plus 0.5em minus 0.4em\relax IEEE, 2016, pp. 4278--4285.

\bibitem{li2021free}
Y.~Li, S.~Deng, X.~Dong, R.~Gong, and S.~Gu, ``A free lunch from ann: Towards
  efficient, accurate spiking neural networks calibration,'' in
  \emph{International Conference on Machine Learning}.\hskip 1em plus 0.5em
  minus 0.4em\relax PMLR, 2021, pp. 6316--6325.

\bibitem{RathiSP020}
\BIBentryALTinterwordspacing
N.~Rathi, G.~Srinivasan, P.~Panda, and K.~Roy, ``Enabling deep spiking neural
  networks with hybrid conversion and spike timing dependent backpropagation,''
  in \emph{8th International Conference on Learning Representations, {ICLR}
  2020, Addis Ababa, Ethiopia, April 26-30, 2020}.\hskip 1em plus 0.5em minus
  0.4em\relax OpenReview.net, 2020. [Online]. Available:
  \url{https://openreview.net/forum?id=B1xSperKvH}
\BIBentrySTDinterwordspacing

\bibitem{KushawahaKBV20}
\BIBentryALTinterwordspacing
R.~K. Kushawaha, S.~Kumar, B.~Banerjee, and R.~Velmurugan, ``Distilling spikes:
  Knowledge distillation in spiking neural networks,'' in \emph{25th
  International Conference on Pattern Recognition, {ICPR} 2020, Virtual Event /
  Milan, Italy, January 10-15, 2021}.\hskip 1em plus 0.5em minus 0.4em\relax
  {IEEE}, 2020, pp. 4536--4543. [Online]. Available:
  \url{https://doi.org/10.1109/ICPR48806.2021.9412147}
\BIBentrySTDinterwordspacing

\bibitem{TakuyaZN21}
\BIBentryALTinterwordspacing
S.~Takuya, R.~Zhang, and Y.~Nakashima, ``Training low-latency spiking neural
  network through knowledge distillation,'' in \emph{{IEEE} Symposium in
  Low-Power and High-Speed Chips, {COOL} {CHIPS} 2021, Tokyo, Japan, April
  14-16, 2021}.\hskip 1em plus 0.5em minus 0.4em\relax {IEEE}, 2021, pp. 1--3.
  [Online]. Available:
  \url{https://doi.org/10.1109/COOLCHIPS52128.2021.9410323}
\BIBentrySTDinterwordspacing

\bibitem{srivastava2014dropout}
N.~Srivastava, G.~Hinton, A.~Krizhevsky, I.~Sutskever, and R.~Salakhutdinov,
  ``Dropout: a simple way to prevent neural networks from overfitting,''
  \emph{The journal of machine learning research}, vol.~15, no.~1, pp.
  1929--1958, 2014.

\bibitem{kim2021visual}
Y.~Kim and P.~Panda, ``Visual explanations from spiking neural networks using
  inter-spike intervals,'' \emph{Scientific reports}, vol.~11, no.~1, pp.
  1--14, 2021.

\bibitem{Parkhi15}
O.~M. Parkhi, A.~Vedaldi, and A.~Zisserman, ``Deep face recognition,'' in
  \emph{British Machine Vision Conference}, 2015.

\bibitem{ioffe2015batch}
S.~Ioffe and C.~Szegedy, ``Batch normalization: Accelerating deep network
  training by reducing internal covariate shift,'' in \emph{International
  conference on machine learning}.\hskip 1em plus 0.5em minus 0.4em\relax PMLR,
  2015, pp. 448--456.

\bibitem{bellec2018long}
G.~Bellec, D.~Salaj, A.~Subramoney, R.~Legenstein, and W.~Maass, ``Long
  short-term memory and learning-to-learn in networks of spiking neurons,''
  \emph{Advances in neural information processing systems}, vol.~31, 2018.

\bibitem{kingma2014adam}
D.~P. Kingma and J.~Ba, ``Adam: A method for stochastic optimization,''
  \emph{arXiv preprint arXiv:1412.6980}, 2014.

\bibitem{mirzadeh2020improved}
S.~I. Mirzadeh, M.~Farajtabar, A.~Li, N.~Levine, A.~Matsukawa, and
  H.~Ghasemzadeh, ``Improved knowledge distillation via teacher assistant,'' in
  \emph{Proceedings of the AAAI conference on artificial intelligence},
  vol.~34, no.~04, 2020, pp. 5191--5198.

\bibitem{rathi2021diet}
N.~Rathi and K.~Roy, ``Diet-snn: A low-latency spiking neural network with
  direct input encoding and leakage and threshold optimization,'' \emph{IEEE
  Transactions on Neural Networks and Learning Systems}, 2021.

\end{thebibliography}

\end{document}